\documentclass[10pt,twocolumn,letterpaper]{article}

\usepackage{cvpr}      
\definecolor{cvprblue}{rgb}{0.21,0.49,0.74}
\usepackage[pagebackref,breaklinks,colorlinks,allcolors=cvprblue]{hyperref}

\def\paperID{18533} 
\def\confName{CVPR}
\def\confYear{2026}

\title{One Latent Space to Rule All Degradations: Unifying Restoration Knowledge for Image Fusion}

\author{Haolong Ma\\
Jiangnan University\\
\and
Hui Li\\
Jiangnan University\\
\and
Chunyang Cheng\\
Jiangnan University
\and
Zeyang Zhang\\
Jiangnan University
\and
Xiaoqing Luo\\
Jiangnan University
\and
Zhongwei Shen\\
Suzhou University of Science and Technology
\and
Xiaoning Song\\
Jiangnan University
\and
Xiao-Jun Wu\\
Jiangnan University
}

\begin{document}
\maketitle
\begin{abstract}
All-in-One Degradation-Aware Fusion Models (ADFMs) as one of multi-modal image fusion models, which aims to address complex scenes by mitigating degradations from source images and generating high-quality fused images. Mainstream ADFMs rely on end-to-end learning and heavily synthesized datasets to achieve degradation awareness and fusion. This rough learning strategy and non-real world scenario dataset dependence often limit their upper-bound performance, leading to low-quality results. To address these limitations, we present LURE, a \textbf{L}earning-driven \textbf{U}nified \textbf{RE}presentation model for infrared and visible image fusion, which is degradation-aware. LURE learns a Unified Latent Feature Space (ULFS) to avoid the dependency on complex data formats inherent in previous end-to-end learning pipelines. It further improves image fusion quality by leveraging the intrinsic relationships between multi-modalities. A novel loss function is also proposed to drive the learning of unified latent representations more stable. 
More importantly, LURE seamlessly incorporates existing high-quality real-world image restoration datasets. To further enhance the model’s representation capability, we design a simple yet effective structure, termed internal residual block, to facilitate the learning of latent features. Experiments show our method outperforms state-of-the-art (SOTA) methods across general fusion, degradation-aware fusion, and downstream tasks. The code is available in the supplementary materials.
\end{abstract}    
\begin{figure}[tb]
  \centering
  \includegraphics[width=0.48\textwidth,height=0.37\textwidth]{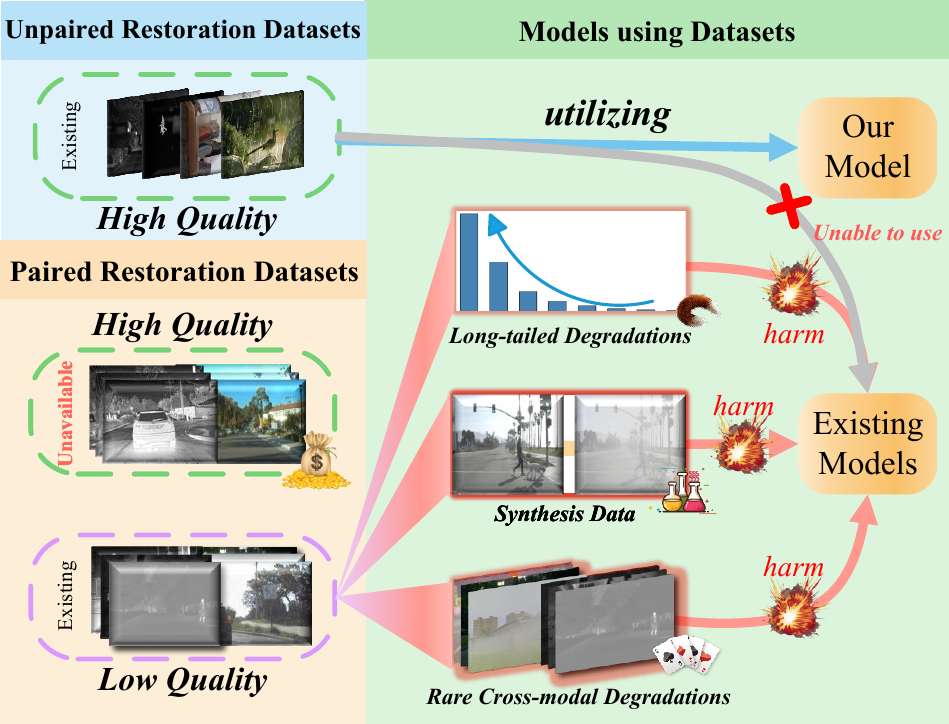}
  \vspace{-1.5\baselineskip}
  \caption{Illustration of the limitations in existing methods due to data-level issues. ``Paired" denotes datasets containing corresponding infrared–visible image pairs.
  }
  \label{fig:first}
\end{figure}
\section{Introduction}
Multi-modal image fusion, aims at creating high-quality fused images by integrating multi-modal sensor data; notably, infrared and visible image fusion is recognized as one of its crucial tasks \cite{bai2025task, karim2023current, zhang2025ddbfusion}. It faces challenges due to degradations in visual information of different modalities. Detailed visible images degrade under varying weather and lighting, causing information loss.  In contrast, infrared images, capturing thermal radiation in adverse environments, typically have lower resolution/contrast.\cite{zhang2021image, yi2024text, liang2022fusion}. Thus, effective degradation removal and complementary data integration are essential for accurate scene reconstruction \cite{cheng2024fusionbooster}. 

Conventional two-step methods first restore and then fuse modalities \cite{yi2024text}, but suffer from domain shift between restoration and fusion. Moreover, diverse degradation combinations (\textit{e.g.}, visible low light and infrared low contrast) make separate fusion training infeasible \cite{li2020mdlatlrr, cheng2025textfusion, zhao2023metafusion}.

Current methods address this problem in one step by training an All-in-One degradation-aware fusion model (ADFM). Mainstream ADFMs typically employ a supervised paradigm, relying on high-quality paired infrared and visible images as explicit supervisory signals to achieve high-quality fusion \cite{yi2024text, conde2024instructir}. This means that each training sample for these methods consists of a quadruplet of four images: two low-quality images of the same scene as input, and two corresponding high-quality images as supervision signals for end-to-end learning \cite{yi2024text} \cite{xu2020u2fusion} \cite{jia2021llvip} \cite{ha2017mfnet}. However, the datasets currently available to support the training of such fusion methods suffer from several critical deficiencies, which in turn constrain the fusion quality of existing models.

As illustrated in Fig.\ref{fig:first}, the infrared and visible modalities exhibit distinct types of degradation. These degradations are combined within the quadruplets required for training. Assuming there are $M$ types of degradation in the infrared modality and $N$ types in the visible modality, a total of $M\times N$ types of quadruplets would be necessary for comprehensive training. This leads to a combinatorial explosion and significantly increases the difficulty of dataset construction. Consequently, existing datasets heavily rely on synthetically generated data to simulate a wide array of degradation scenarios  \cite{xu2020u2fusion} \cite{jia2021llvip} \cite{ha2017mfnet}.

For existing IVIF datasets (\textit{e.g.}, RoadScene \cite{xu2020aaai}, MSRS \cite{Tang2022PIAFusion}, M3FD \cite{liu2022target}), degraded images are typically handled through two strategies. The forward construction method restores degraded images using SOTA restoration techniques or manual adjustment to obtain ground truth, while the backward construction method artificially adds degradations to clean images when real degraded samples are scarce. Forward construction is limited by the scarcity of real degraded scenes. In contrast, backward construction often introduces a distribution gap from real degradations—especially for rare cases. Current datasets follow a long-tail distribution, with most samples covering common degradations (e.g., low light, overexposure) and only a few representing rare but critical ones (e.g., dense fog).

To address these issues, we propose a representation learning method based on decoupling. First, in the initial stage, we leverage existing high-quality image restoration datasets for each modality independently. The objective is to learn a unified latent representation space that is agnostic to the specific type of degradation. In this space, the original degradation type cannot be inferred from an image’s representation, making it degradation-agnostic. It serves as a unified representation where all degraded images are mapped, effectively eliminating the influence of specific degradations. In the second stage, we build upon the features from this representation space to learn how to generate high-quality fused results—that is, to learn the fusion rules. Since all types of degradation are mapped to a common intermediate form, a single training process is sufficient to handle input images with any form of degradation. Importantly, for the first stage, we only require image restoration datasets, some of which provide high-quality supervision from real-world scenes (\textit{e.g.}, RESIDE \cite{li2019benchmarking}, LOL \cite{wei2018deep}). For the second stage, we only need a standard image fusion dataset. This two-stage design enables us to circumvent the limitations of existing methods.

However, learning an effective image representation requires us to reconstruct the input image, ensuring that its representation preserves the critical information of the image. Existing image restoration models typically employ a skip connection \cite{chen2022simple} \cite{zamir2022restormer} \cite{li2022all} \cite{chen2023comparative}. from input to output to accelerate model convergence, allowing the model to focus on the degraded regions of the image. While this architecture hinders the learning of a comprehensive image representation, removing it leads to slower convergence. Therefore, we adapt this residual learning mechanism by shifting from the conventional pixel-level residual to a feature-level residual, proposing an ``inner residual structure" to resolve the aforementioned issue.
The contributions of this paper can be summarized as follows:

- A novel data-decoupling and feature association method is proposed to address challenges in synthetic data over-reliance and jointly handle cross-model combined degradations.

- An efficient inner residual structure is introduced to enhance the model's information retention while ensuring its ability to learn complete latent features.

- Extensive experiments are conducted. Results demonstrate the effectiveness of the proposed approach.
\section{Related Work}
\subsection{Recent Image Fusion Methods}
Mainstream image fusion models fall into three categories: generative models (\textit{e.g.}, FusionGAN, DDFM)\cite{zhao2023ddfm, ma2020ganmcc, ma2019fusiongan}, end-to-end approaches (\textit{e.g.}, U2Fusion)\cite{li2025conti, wang2025highlight, zhao2023cddfuse}, and autoencoder-based methods (\textit{e.g.}, DenseFuse)\cite{li2018densefuse, li2020nestfuse, li2024crossfuse}. Generative models fuse images via distribution approximation, yet face training instability and heuristic designs. End-to-end methods directly derive fused images, constrained by designed loss functions.  

Despite stable training, their generalization is limited by handcrafted loss functions. Early autoencoders lacked adaptability with hand-engineered fusion strategies\cite{li2018densefuse}, whereas current methods use learnable rules for better adaptation \cite{li2024crossfuse}. Autoencoder methods inherently decouple fusion rule and feature learning, facilitating versatile, degradation-aware paradigms.

\subsection{All-in-One Degradation-aware Models}

All-in-One Degradation-aware Models (ADM) address diverse degradations within a single model\cite{conde2024instructir, li2024promptcir, li2022all}.  In image restoration field,  ADMs benefit from abundant high-quality datasets and use techniques like prompt/continual learning\cite{wei2018deep, li2019benchmarking, afifi2021learning}. 

However, in image fusion field, ADFMs face challenges due to long-tailed degradation distributions in fusion datasets, lacking sufficient data for robust degradation restoration\cite{yi2024text, liu2022target, xu2020aaai, Tang2022PIAFusion}. Unsupervised methods like contrastive learning exist but often yield lower quality due to lacking explicit supervision\cite{wang2025degradation}. Current ADFMs still rely on existing fusion dataset degradations or synthetic ones, which may not match real degradation distributions\cite{yi2024text, wang2025degradation}. Some degradations require extra modality information for realistic synthesis (\textit{e.g.}, fog needs depth maps), which is unavailable in current datasets.

These limitations hinder the generalization of current ADFMs. To address the above drawbacks, we propose a unified degradation-aware representation model to decouple image restoration and fusion data-level and re-associate them in latent space, which allows leveraging high-quality image restoration data to overcome current ADFMs challenges.
\section{Proposed Method}
\subsection{Problem Formulation}


In this paper, low/high-quality images are denoted by sets $\mathcal{X}$/$\mathcal{Y}$. Infrared/visible images are denoted by sets $\mathcal{I}$/$\mathcal{V}$.

Assume the set $\mathcal{X}$ of $T$ degradation types. For the $t$-th degradation ($t \in \{1, 2, ..., T\}$), let $c_t$ denote its description (\textit{e.g.}, text embedding), with $c_t \in \mathcal{C}$ and $\mathcal{C}= \{ c_t \}_{t=1}^{T}$. Theoretically, for each image $x \in \mathcal{X}$, there exists a deterministic mapping to a corresponding degradation description $c_t$.

For supervised method, samples are quadruplets $(x_{ir}, y_{ir}, x_{vi}, y_{vi})$, where $ir$/$vi$ denote infrared/visible modalities, and $x$/$y$ are low/high-quality images. Define modality-paired sets: $\mathcal{I}_g$ (infrared), $\mathcal{V}_g$ (visible):
\begin{equation} 
\mathcal{I}_g = \{ (x_{ir}, y_{ir}) \mid x_{ir} \in \mathcal{X}, y_{ir} \in \mathcal{Y}, y_{ir} = f(x_{ir}) \}, \end{equation}
where $f$ maps low-quality images to clean counterparts. $\mathcal{V}_g$ is defined analogously.

Given a sample $(x_{ir}, y_{ir}, x_{vi}, y_{vi})$ from the joint distribution $P_{\mathcal{I}_g, \mathcal{V}_g}(x_{ir}, y_{ir}, x_{vi}, y_{vi})$ and the description $c$, the loss function of existing supervised methods can be expressed as $\mathcal{L}_{if}(y_{ir}, y_{vi}, \mathcal{F}_{if}(x_{ir}, x_{vi}, c; \theta))$
where, $\mathcal{L}_{if}$ is the fusion loss. $\mathcal{F}_{if}$ is a network parameterized by $\theta$, fusing degraded infrared and visible images with degradation description $c$.

Such supervised approaches necessitate quadruplet samples from the joint distribution of $\mathcal{I}_g$ and $\mathcal{V}_g$ to provide supervision signals. This leads to the curse of dimensionality in handling combined degradations, fundamentally due to coupling modalities and quality dimensions at the data level.

\subsection{Data Decoupling and Feature Association}
\begin{figure}[!h]
  \includegraphics[width=0.48\textwidth, height=0.27\textwidth]{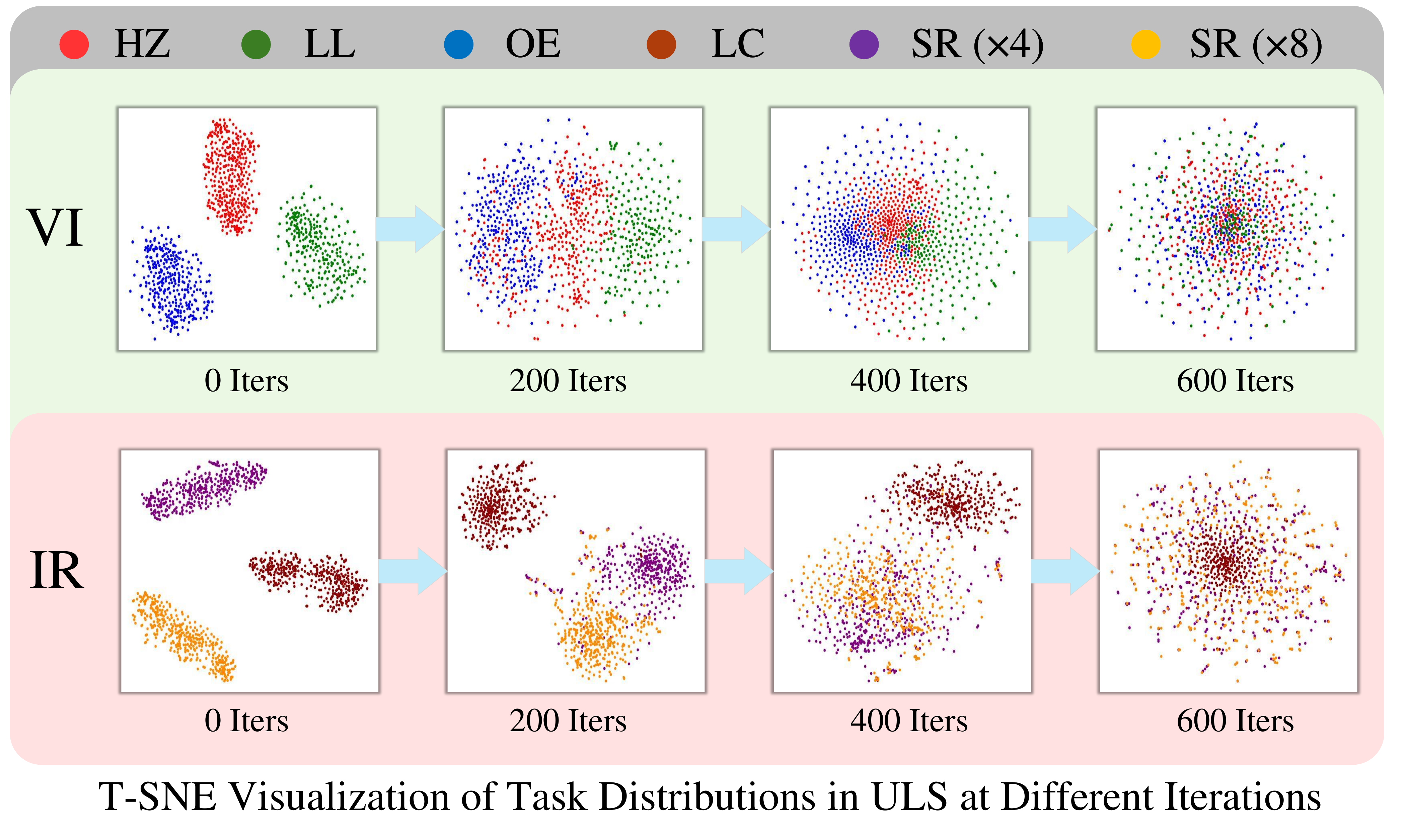}
  \vspace{-1.5\baselineskip}
  \caption{
    t-SNE visualization of ULFS distributions for tasks (\textit{e.g.,} HZ denotes Dehaze; other abbreviations are detailed in Sec.\ref{sec:data}.) of both modalities across training iterations. It reveals initial distinct task distributions gradually merging into a unified distribution.  Detailed t-SNE visualizations are in the supplementary material.
  }
  \label{fig:visual}
\end{figure}

To address modality-quality coupling in data-level, a novel decoupling mechanism which still preserves inherent associations of source data is needed. We thus propose a ULFS ($\mathcal{Z}$) to re-establish these associations at feature level. We define $\mathcal{Z}$ as follows:

\textbf{Definition: [Unified Latent Feature Space ($\mathcal{Z}$)].}  
\begin{quote}
$\mathcal{Z}$ is defined by mapping $f_e: \mathcal{X} \to \mathcal{Z}$, with conditions:
\begin{enumerate}[label=(\Roman*)]
    \item $\forall x \in \mathcal{X}, \exists z \in \mathcal{Z}$ ($z = f_e(x))$.
    \item $\forall z \in \mathcal{Z}, \forall c \in \mathcal{C}$ ($P(c \mid z) = P(c))$.
\end{enumerate}
\end{quote}
\begin{enumerate}
    \item Condition \uppercase\expandafter{\romannumeral 1} ensures every degraded image maps to $\mathcal{Z}$. 
    \item Condition \uppercase\expandafter{\romannumeral 2} ensures that for any latent feature $z \in \mathcal{Z}$, the posterior $P(c|z)$ distribution equals the prior $P(c)$ distribution for $c \in \mathcal{C}$.
\end{enumerate}

Practically, each modality learns its own $\mathcal{Z}$ for quality decoupling. These $\mathcal{Z}$ properties guarantee $f_e$ mapping eliminates unique degradation information. As shown in Fig.\ref{fig:visual}, this `mixing' of feature distributions in $\mathcal{Z}$ renders degradations indistinguishable, allowing transfer of fusion rules across degradations. 

We learn $\mathcal{Z}$ for image quality using image restoration data at first stage. Then, at second stage, we learn a fusion fule in $\mathcal{Z}$ using image fusion data.

\subsubsection{Unified Latent Representation Learning}
\begin{figure*}[tb]
  \includegraphics[width=1\textwidth]{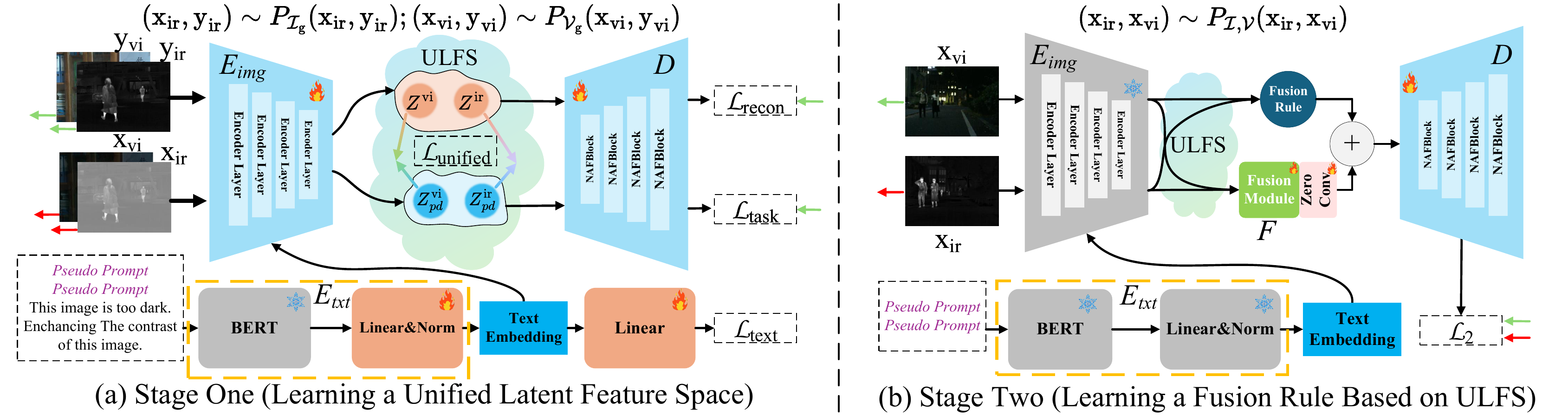}
  \vspace{-1.5\baselineskip}
  \caption{Two-stage training overview. (a) Stage I learns a unified latent space guided by text. (b) Stage II freezes encoders and trains a fusion module for strategy learning. “Pseudo prompt” denotes the prompt for pseudo degradation (details in Supplementary).
  } 
  \label{fig:training}
\end{figure*}
Theoretically, to align with the definition of $\mathcal{Z}$ for ULFS learning, we aim to minimize the distribution distance of degraded images within $\mathcal{Z}$. Learning Objective:
\begin{equation}
\label{eq:kl}
\min_{\theta} \sum_{t=1}^T \sum_{t' \neq t}^{T} \text{KL}(P^{(t)}_{\mathcal{Z}}(z) \parallel P^{(t')}_{\mathcal{Z}}(z)),
\end{equation}
where, $\text{KL}(\cdot \parallel \cdot)$ denotes the Kullback-Leibler divergence, and $P^{(t)}_{\mathcal{Z}}(z)$ represents the distribution of the $t$-th degradation in $\mathcal{Z}$.

To minimize Eq.\ref{eq:kl}, Generative Adversarial Networks (GANs) with discriminators are commonly used for adversarial training \cite{li2023learning, poirier2023robust}.  However, inherent distribution differences exist across image restoration datasets. As the discriminators tend to capture these discrepancies, the GAN training process becomes unstable and struggles to converge.

To mitigate this issue, we introduce a pseudo-degradation task.  This task, fundamentally image reconstruction, is not a genuine degradation. For this task, pseudo-degradation data pairs $(x, y)$ satisfy $x = y$, where $x \in \mathcal{X}$ and $y \in \mathcal{Y}$. We control this task via a description $c_{\mathrm{pd}}$ and align all degradation distributions to this pseudo-degradation task distribution.

For the $t$-th degradation, given an image restoration data pair $(x_{t}, y_{t})$ where $x_{t} \in \mathcal{X}$ and $y_{t} \in \mathcal{Y}$, we have $z_{t} = f_e(x_t, c_t)$. We construct a pseudo-task data pair $(y_t, y_t)$, yielding $z_{\mathrm{pd}} = f_e(y_t, c_{\mathrm{pd}})$.

Theoretically, $z_{\mathrm{pd}}$ should ideally correspond to $z_t$ as a feature point in ULFS. This is justified by:

1. $x_t$ and $y_t$ represent the same scene information.

2. Pseudo-degradation, being image reconstruction of degradation-free $y_t$, results in $z_{\mathrm{pd}} \in \mathcal{Z}$ lacking degradation information, thus obscuring original degradation types.

Therefore, to align distributions of tasks in $\mathcal{Z}$, we ensure each task's representation aligns with its pseudo-task counterpart.  We transform distribution alignment in the objective function to feature alignment:
\begin{equation}
\mathcal{L_{\mathrm{unified}}} = \frac{1}{T} \sum_{t=1}^T \{1 - \Gamma[f_e(X_t, c_t), f_e(Y_t, c_{\mathrm{pd}})]\},
\end{equation}
where, $X_t$ and $Y_t$ represent low-quality and high-quality image datasets for task $t$, respectively. $\Gamma$ measures feature distances in $\mathcal{Z}$, employing cosine similarity.

It is worth emphasizing that, in practice, the distributional discrepancies among different datasets still prevent Condition II in the ULFS definition from being perfectly satisfied. Consequently, the learned latent space can only approximately meet this condition. Nevertheless, directly constraining feature relationships for $\mathcal{Z}$ learning proves to be more stable than GAN-based adversarial approaches. The pseudo-task acts as an image auto-encoding process to learn the fusion rule in the second stage and to bridge the two stages at the feature level. In practice, $f_e$ functions as a conditional image encoder.

\subsection{Network Pipeline}

\subsubsection{Encoder and Decoder Structure}

As shown in Fig.\ref{fig:training} (a), the encoder is bifurcated into image ($E_{\mathrm{img}}$) and text ($E_{\mathrm{txt}}$) encoders.  $E_{\mathrm{txt}}$ employs Distilled BERT \cite{devlin2019bert}, with a Norm layer and a Linear layer to project its final output into a text feature vector.  A classification head and cross-entropy loss ($\mathcal{L}_{\mathrm{text}}$) are used to align task categories for each task-specific text feature vector, an approach effective in InstructIR \cite{oh2024instructir} and superior to the CLIP text encoder \cite{radford2021learning}. 

\begin{figure}[!ht]
  \includegraphics[width=0.48\textwidth, height=0.15\textwidth]{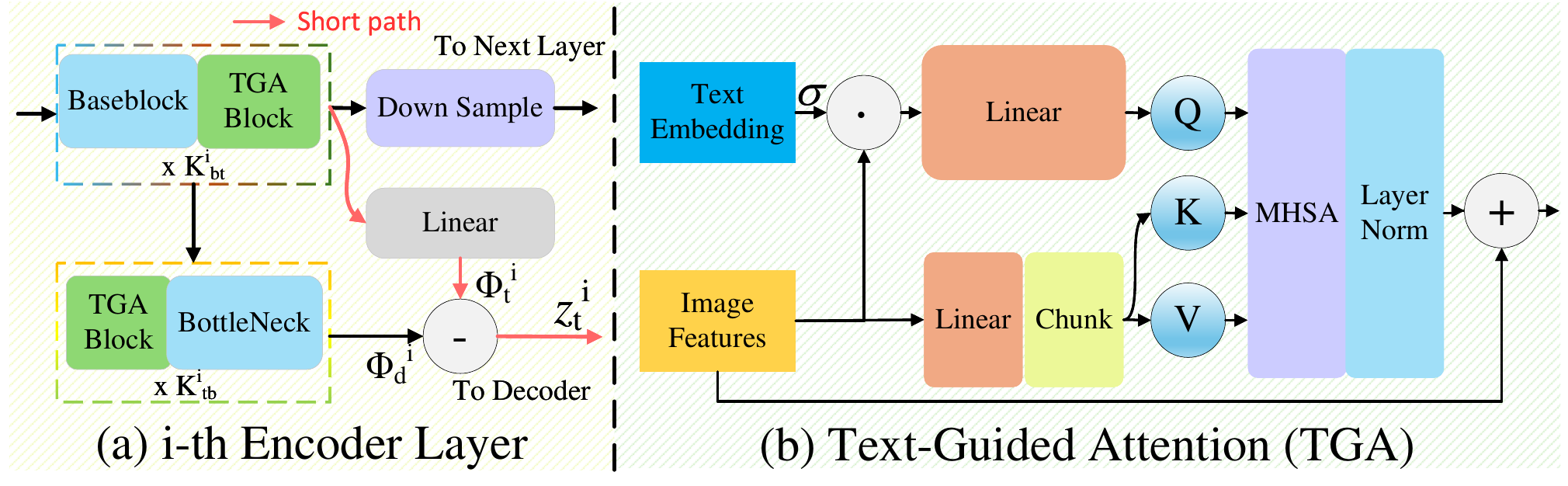}
  \vspace{-1.5\baselineskip}
  \caption{
    Schematic diagrams of Encoder Layer and Text-Guided Attention (TGA). (a) Structure diagram of the Encoder Layer. (b) Structure diagram of TGA. ``$\odot, \oplus, \ominus$'' represent Element-wise operations, and $\sigma$ represents the Sigmoid function.
  }
  \label{fig:encoder}
\end{figure}

$E_{\mathrm{img}}$ comprises multi-scale Encoder Layers for mapping and reconstructing input images to learn $\mathcal{Z}$. Unlike conventional restoration models using residual structures ($\hat{y}=\mathcal{M}(x)+x$) \cite{zamir2022restormer, li2022all, chen2022simple}, which emphasize degraded-region correction over source reconstruction, our framework requires balanced feature learning. To avoid detail loss from removing the residual path, we introduce an inner residual structure to preserve fine information. Each Encoder Layer (Fig.\ref{fig:encoder}) includes four modules—BaseBlock, TGABlock (Text-Guided Attention Block), BottleNeck, and Linear—where TGABlock is integrated with BaseBlock and BottleNeck to form two composite units, iterated $K^{i}_{tb}$ and $K^{i}_{tb}$ times in the $i$-th layer.

As shown in Fig.\ref{fig:encoder}, at the $i$-th layer, TGABlock + BaseBlock extract modality-specific features and integrate degradation information guided by text. The output branches into three paths: (1) downsampling to the next layer, (2) TGABlock + BottleNeck for degradation-specific feature extraction $\Phi_d^{i}$, and (3) a linear layer for task-relevant base information $\Phi_t^{i}$. A residual subtraction yields task-irrelevant latent representations $z_t^i = \Phi_t^{i} - \Phi_d^{i}$, providing a shorter decoding path and mitigating detail loss.

BaseBlock (NAFBlock \cite{chen2022simple} + Modality Embedding) extracts modality-aware basic features, while BottleNeck (NAFBlock + $1\times1$ conv) retains text-relevant cues and reduces redundancy. The TGABlock adopts a Transformer-like design with GDFN \cite{zamir2022restormer} as its feed-forward block. Unlike passive channel-weighting schemes \cite{yi2024text, oh2024instructir}, TGABlock employs channel-weighted text features as Queries to actively attend to global image features, enhancing text-guided spatial awareness and preserving task-related details.

Each decoder layer then upsamples and refines details from $\mathcal{Z}$ using a few NAFBlocks, while the fusion stage incorporates multi-scale Cross Attention modules for effective multimodal interaction.

\subsubsection{Training}
In our proposed framework, the training processing is a two-stage scheme. As shown in Fig.\ref{fig:training} (a), in the first stage, the latent space $\mathcal{Z}$ is learned. For each sample, we have $(x_{ir}, y_{ir})\sim P_{\mathcal{I}_g}(x_{ir}, y_{ir})$ or $(x_{vi}, y_{vi})\sim P_{\mathcal{V}_g}(x_{vi}, y_{vi})$. To explicitly represent modality information, we transform the data into triplets $(x, y, m)$, where $m \in \{0, 1\}$; 0 denotes visible modality, and 1 denotes infrared modality.

As shown in Fig.\ref{fig:training} (b), stage two learns fusion rules. Input pairs are $(x_{ir}, x_{vi})$ from image fusion datasets: $(x_{ir}, x_{vi})\sim P_{\mathcal{I},\mathcal{V}}(x_{ir}, x_{vi})$.

Stage one uses image restoration datasets, stage two uses an image fusion dataset.

\textbf{First stage training}: In first stage, the unified latent space, $\mathcal{Z}$, is learned. Image and text encoders extract multi-scale features for both real and pseudo-task data. Given low-quality image $x$, high-quality image $y$, text $\omega$, description $c_t=E_{\mathrm{txt}}(\omega)$, pseudo-task text $\omega_{pd}$, pseudo description $c_{pd}=E_{\mathrm{txt}}(\omega_{pd})$, and modality $m$. We extract latent features: $
Z=E_{\mathrm{img}}(x, c_t,m)$ and $
Z_{pd}=E_{\mathrm{img}}(y, c_{pd},m)$
where $Z, Z_{\mathrm{pd}}$ are multi-scale latent representations. The decoder reconstructs images: $\hat y=D(Z)$ and $\hat y_{pd}=D(Z_{pd})$.

The first stage loss $\mathcal{L}_1$ is formulated as follows:
\begin{equation}
\mathcal{L}_1=\mathcal{L}_{\mathrm{task}}+\alpha_1^{(1)} \mathcal{L}_{\mathrm{unified}} + \alpha_2^{(1)} \mathcal{L}_{\mathrm{recon}}+\alpha_3^{(1)}\mathcal{L}_{\mathrm{text}},
\end{equation}
where
\begin{enumerate}
\item $\mathcal{L}_{\mathrm{recon}}=||\hat y_{pd}-y||_1$ represents the reconstruction loss for the pseudo-degradation task.
\item $\mathcal{L}_{\mathrm{task}}=||\hat y-y||_1+|||\nabla \hat y|-|\nabla y|||_1$ represents the loss for different degradation tasks ($\nabla$: Sobel operator)
\item $\mathcal{L}_{\mathrm{unified}}=1 - \Gamma(Z,Z_{\mathrm{pd}})$, ($\Gamma$: Cosine Simlarity)
\item $\alpha_1^{(1)}, \alpha_2^{(1)}, \alpha_3^{(1)}$ are hyperparameters.
\end{enumerate}

\textbf{Second stage training}: In second stage, the proposed model learns fusion rules based on $\mathcal{Z}$. Given an infrared image $x_{ir}$, visible image $ x_{vi}$, the pseudo-degradation prompt $\omega_{pd}$ and pseudo-degradation task description ($c_{pd}=E_{\mathrm{txt}}(\omega_{pd})$), we extract latent
features:
\begin{equation}
Z_{\mathrm{pd}}^\mathrm{ir}=E_{\mathrm{img}}(x_{ir},c_\mathrm{pd},1), \quad Z_{\mathrm{pd}}^\mathrm{vi}=E_{\mathrm{img}}(x_{vi},c_\mathrm{pd},0).
\end{equation}
Fusion result $O_{\mathrm{fused}}$ is obtained via:
\begin{gather}
O_{\mathrm{fused}}=D\{zero[F(Z_{\mathrm{pd}}^\mathrm{vi},Z_{\mathrm{pd}}^\mathrm{ir})]+rule(Z_{\mathrm{pd}}^\mathrm{vi},Z_{\mathrm{pd}}^\mathrm{ir})\}.
\end{gather}

Inspired by ControlNet \cite{zhang2023adding}, we first implement a prior fusion rule, $rule(\cdot,\cdot)$, for initial feature fusion.  A learnable module, $F(\cdot,\cdot)$, then refines rule-based fusion results.  A zero-initialized $1\times 1$ convolution mitigates random initialization noise and accelerates Second Stage convergence. 

The Second Stage loss function is defined as follows:
\begin{equation}\mathcal{L}_2=\mathcal{L}_\mathrm{color}+\alpha_1^{(2)}\mathcal{L}_\mathrm{grad}+\alpha_2^{(2)}\mathcal{L}_\mathrm{per},
\end{equation}
where $\alpha_1^{(2)}$ and $\alpha_2^{(2)}$ are hyperparameters balancing loss contributions. 
Since the proposed unified latent space can extract powerful degradation irrelevant features, theoretically, the fusion loss function not need to be specially designed. Thus, the color consistency item ($\mathcal{L}_\mathrm{color}$) and gradient item ($\mathcal{L}_\mathrm{grad}$) are chosen from Text-IF \cite{yi2024text}. The perceptual item ($\mathcal{L}_\mathrm{per}$) is chosen from U2Fusion \cite{xu2020u2fusion}.

\subsubsection{Inference}
For inference with infrared and visible images ($x_{ir}, x_{vi}$) and degradation description $c_{t}=E_{\mathrm{txt}}(\omega)$ and pseudo-degradation $c_{pd}=E_{\mathrm{txt}}(\omega_{pd})$ (assuming infrared is degraded), the process of LURE is given as follows:
\begin{equation}Z_{\mathrm{pd}}^\mathrm{vi}=E_{\mathrm{img}}(x_{vi},c_{pd},0), \quad Z_{\mathrm{t}}^\mathrm{ir}=E_{\mathrm{img}}(x_{ir},c_{t},1),
\end{equation}
\begin{gather}O_{\mathrm{fused}}=D\{zero[F(Z_{\mathrm{pd}}^\mathrm{vi},Z_{\mathrm{t}}^\mathrm{ir})]+rule(Z_{\mathrm{pd}}^\mathrm{vi},Z_{\mathrm{t}}^\mathrm{ir})\}.
\end{gather}

End-to-end inference uses one model, where $c_{\textrm{pd}}$ and $c_t$ denote non-degraded and degraded modalities. This design enables diverse and combined degradation handling via descriptive conditions, without requiring joint degradation training data.

\begin{figure*}[!ht]
  \includegraphics[width=1\textwidth, height=0.48
\linewidth]{6cmp.pdf}
  \vspace{-1.6\baselineskip}
  \caption{Qualitative comparison of degradation-aware fusion tasks. ``eir" denotes External Image Restoration Methods. Text-IF uses ``eir" only for Super Resolution and combined degradations. For more detail about the qualitative comparison, please refer to supplementary materials.
  }
  \label{fig:qualitative}
\end{figure*}

\section{Experiment}
\subsection{Setting}
In the proposed model, the encoder and the decoder both consist of 4 layers. The channel numbers for each layer are [48, 96, 192, 384], reduced to [16, 32, 64, 128] for each encoder layer's output size. 
$K_{bt}$ and $K_{tb}$ are set to [1,1,2,2] and [2,2,4,8], respectively. Decoder blocks and fusion blocks for each layer is set to [1,1,1,1] and [1, 1, 2, 2]. The number of attention heads for TGA is set to 4, and the prior rule is selected from DenseFuse \cite{li2018densefuse}. \footnote{\label{ablation}More detailed hyperparameter (\textit{e.g.} hyperparameter ablation) and training settings are provided in the supplementary materials}

\subsubsection{Training Data}
\label{sec:data}
For the first stage of training, we utilize the most common tasks for infrared modality: low contrast (LC) and low resolution (SR) ($\times$4 and $\times$8).  We use the EMS Full dataset \cite{yi2024text} and construct a high-quality super-resolution dataset on LLVIP \cite{jia2021llvip}, following the approach in InstructIR \cite{oh2024instructir}. For visible modality, we perform dehazing (HZ), overexposure correction (OE), and low-light enhancement (LL). The high-quality images are chosen from the RESIDE \cite{li2019benchmarking}, Exposure-Errors \cite{afifi2021learning}, and LOL datasets \cite{wei2018deep}, respectively. Our method does not entirely exclude synthetic datasets but reduces reliance on low-quality ones, using real-world and high-quality synthetic restoration data.

For the second stage, the training data of the MSRS datasets \cite{Tang2022PIAFusion} is used. All of the aforementioned datasets are publicly available. 

All task texts are generated by LLaMA \cite{touvron2023llama}.

\subsubsection{Evaluation Data}

For vanilla image fusion, we evaluate the fusion performance on the RoadScene \cite{xu2020aaai}, M3FD \cite{Tang2022PIAFusion}, TNO \cite{tno}, and MSRS \cite{Tang2022PIAFusion} test dataset. 

For degradation-aware image fusion, we use EMS-full \cite{yi2024text} for low contrast, overexposure correction, and low-light enhancement. We also construct a super-resolution test dataset on M3FD \cite{Tang2022PIAFusion} in the same manner. For dehazing, we collect hazy samples from M3FD \cite{Tang2022PIAFusion} to create a real-world dehazing test dataset. 

For cross-modal combined degradations, we exemplify with low contrast + overexposure correction and low contrast + low light enhancement, creating combined test datasets from simgle modality degradation tasks. 

\subsubsection{Evaluation Metrics}
The evaluation metrics include Correlation Coefficient (CC) \cite{deshmukh2010image}, Spatial Correlation of Differences (SC) \cite{aslantas2015new}, Structural Similarity Index (SS) \cite{wang2002universal}, Peak Signal-to-Noise Ratio (PS), Multiscale Structural Similarity (MS) \cite{wang2003multiscale}, Average Gradient (AG), Standard Deviation (SD), Spatial Frequency (SF), Entropy (EN) \cite{roberts2008assessment}, CLIP-IQA (CL) \cite{wang2023exploring}, LIQE (LI) \cite{zhang2023liqe}, MANIQA (MA) \cite{yang2022maniqa}, NIQE (NI) \cite{mittal2012making}, NUSIQ \cite{ke2021musiq} (NU), and ARNIQA (AR) \cite{agnolucci2024arniqa}. 

These metrics comprehensively evaluate the fused image quality from multiple perspectives, including image noise, texture details, and preservation of source image structure. CLIP-IQA \cite{wang2023exploring}, LIQE \cite{wang2023exploring}, MANIQA \cite{yang2022maniqa}, NUSIQ \cite{ke2021musiq}, and ARNIQA \cite{agnolucci2024arniqa} are state-of-the-art no-reference image quality assessment metrics. Except for NIQE, higher values indicate better fusion quality. Red and blue values indicate best and second best results.

\subsection{SOTA Competitors}
We conducted comparisons against seven state-of-the-art (SOTA) methods, including LRRNet (LRRN) \cite{li2023lrrnet}, TITA (TITA) \cite{hu2025balancing}, SAGE (SAGE) \cite{wu2025every}, MMDRFuse (MMDR) \cite{deng2024mmdrfuse}, Text-IF (Text) \cite{yi2024text}, DAFusion (DAFu) \cite{wang2025degradation}, and CDDFuse (CDDF) \cite{zhao2023cddfuse}. Among these, Text-IF and DAFusion are categorized as All-in-One degradation-aware image fusion models. It should be noted that DAFusion (DAFu) is not included for fairness, since it inherently performs restoration on all inputs.

\subsection{Vanilla Image Fusion}

\begin{table}[!ht]
\renewcommand{\arraystretch}{0.7}
\setlength{\tabcolsep}{1.3mm}{
\caption{Quantitative comparison of vanilla image fusion task on MSRS and M3FD.}
\label{tab:vanilla_cmp_1}
\vspace{-0.3cm}
{\fontsize{7pt}{12pt}\selectfont
\begin{tabular}{c|ccccc|ccccc}
\hline
 & \multicolumn{5}{c|}{RoadScene Dataset} & \multicolumn{5}{c}{MSRS Dataset} \\ \hline
Methods & EN & SC & PS & SS & MS & EN & SC & PS & SS & MS \\ \hline
LRRN'23 & {\color[HTML]{3166FF} \textbf{7.13}} & 1.57 & 59.94 & 0.34 & 0.49 & 6.19 & 0.79 & 64.74 & 0.21 & 0.37 \\
CDDF'23 & {\color[HTML]{FE0000} \textbf{7.45}} & {\color[HTML]{FE0000} \textbf{1.71}} & 62.14 & 0.48 & {\color[HTML]{3166FF} \textbf{0.53}} & {\color[HTML]{FE0000} \textbf{6.70}} & 1.62 & 64.41 & {\color[HTML]{FE0000} \textbf{0.50}} & 0.51 \\
MMDR'24 & 6.51 & 1.10 & 62.55 & 0.41 & 0.48 & 6.48 & 1.53 & 64.95 & 0.48 & 0.52 \\
Text'24 & 6.64 & 1.57 & 62.43 & 0.48 & 0.52 & 6.65 & {\color[HTML]{3166FF} \textbf{1.68}} & 64.59 & 0.48 & 0.52 \\
SAGE'25 & 7.02 & 1.52 & 60.85 & 0.42 & 0.51 & 6.00 & 1.42 & {\color[HTML]{FE0000} \textbf{65.28}} & 0.45 & 0.48 \\
TITA'25 & 7.09 & 1.30 & {\color[HTML]{3166FF} \textbf{63.02}} & {\color[HTML]{3166FF} \textbf{0.49}} & 0.51 & 6.57 & 1.61 & 65.01 & 0.49 & {\color[HTML]{3166FF} \textbf{0.52}} \\ \hline
Ours & {\color[HTML]{FE0000} \textbf{7.45}} & {\color[HTML]{3166FF} \textbf{1.67}} & {\color[HTML]{FE0000} \textbf{63.04}} & {\color[HTML]{FE0000} \textbf{0.50}} & {\color[HTML]{FE0000} \textbf{0.55}} & {\color[HTML]{3166FF} \textbf{6.67}} & {\color[HTML]{FE0000} \textbf{1.74}} & {\color[HTML]{3166FF} \textbf{65.12}} & {\color[HTML]{3166FF} \textbf{0.49}} & {\color[HTML]{FE0000} \textbf{0.53}} \\ \hline
\end{tabular}
}
}
\end{table}

\begin{table}[!ht]
\renewcommand{\arraystretch}{0.7}
\setlength{\tabcolsep}{1.3mm}{
\caption{Quantitative comparison of vanilla image fusion task on TNO and RoadScene.}
\label{tab:vanilla_cmp_2}
\vspace{-0.3cm}
{\fontsize{7pt}{12pt}\selectfont
\begin{tabular}{c|lllll|lllll}
\hline
 & \multicolumn{5}{c|}{TNO Dataset} & \multicolumn{5}{c}{M3FD Dataset} \\ \hline
Methods & \multicolumn{1}{c}{EN} & \multicolumn{1}{c}{SC} & \multicolumn{1}{c}{PS} & \multicolumn{1}{c}{SS} & \multicolumn{1}{c|}{MS} & \multicolumn{1}{c}{EN} & \multicolumn{1}{c}{SC} & \multicolumn{1}{c}{PS} & \multicolumn{1}{c}{SS} & \multicolumn{1}{c}{MS} \\ \hline
LRRN'23 & 6.84 & 1.62 & 62.66 & 0.40 & 0.43 & 6.33 & 1.38 & {\color[HTML]{3166FF} \textbf{64.14}} & 0.39 & 0.43 \\
CDDF'23 & {\color[HTML]{3166FF} \textbf{6.94}} & 1.79 & 62.43 & {\color[HTML]{3166FF} \textbf{0.50}} & {\color[HTML]{3166FF} \textbf{0.46}} & {\color[HTML]{3166FF} \textbf{6.75}} & 1.59 & 62.94 & {\color[HTML]{FE0000} \textbf{0.52}} & 0.52 \\
MMDR'24 & 6.31 & 1.59 & 62.37 & 0.48 & 0.43 & 6.33 & 1.40 & 62.27 & 0.49 & 0.50 \\
Text'24 & {\color[HTML]{FE0000} \textbf{7.04}} & 1.67 & 62.59 & 0.47 & 0.44 & 6.71 & 1.41 & 63.35 & 0.48 & 0.48 \\
SAGE'25 & 6.77 & {\color[HTML]{3166FF} \textbf{1.79}} & 62.00 & 0.49 & 0.45 & {\color[HTML]{3166FF} \textbf{6.74}} & {\color[HTML]{3166FF} \textbf{1.66}} & 62.69 & 0.50 & {\color[HTML]{3166FF} \textbf{0.52}} \\
TITA'25 & 6.70 & 1.62 & {\color[HTML]{3166FF} \textbf{62.71}} & {\color[HTML]{FE0000} \textbf{0.51}} & 0.43 & 6.53 & 1.33 & 63.75 & {\color[HTML]{3166FF} \textbf{0.51}} & 0.50 \\ \hline
Ours & 6.72 & {\color[HTML]{FE0000} \textbf{1.84}} & {\color[HTML]{FE0000} \textbf{63.84}} & {\color[HTML]{FE0000} \textbf{0.51}} & {\color[HTML]{FE0000} \textbf{0.47}} & 6.67 & {\color[HTML]{FE0000} \textbf{1.67}} & {\color[HTML]{FE0000} \textbf{64.24}} & 0.50 & {\color[HTML]{FE0000} \textbf{0.53}} \\ \hline
\end{tabular}
}
}
\end{table}
To ensure fair comparison, we first tested degradation-unaware fusion tasks.  
Quantitative results are in Tab.\ref{tab:vanilla_cmp_1} and Tab.\ref{tab:vanilla_cmp_2}. Our method achieves SOTA across all datasets, demonstrating superior performance. Leveraging high-quality image restoration datasets, our method attains higher PS (PSNR), indicating less noisy fusion results.  Furthermore, Inner Residuals help preserve more source information, leading to better CC and SC (SCD), and improved SS/MS (SSIM/MS-SSIM), signifying better source information retention and structural fidelity at multiple scales.\footnote{For qualitative evaluation results, due to page limitations, please refer to the supplementary materials.}

\subsection{Degradation-aware Image Fusion}

\begin{table}[h]
\renewcommand{\arraystretch}{0.7}
\setlength{\tabcolsep}{0.8mm}{
\caption{Quantitative comparison of degradation-Aware fusion tasks. $\dagger$ denotes All-in-One type methods.}
\label{tab:deg_cmp_1}
\vspace{-0.3cm}
{\fontsize{7pt}{12pt}\selectfont
\begin{tabular}{c|ccc|ccc|ccc|ccc}
\hline
 & \multicolumn{3}{c|}{LL} & \multicolumn{3}{c|}{HZ} & \multicolumn{3}{c|}{OE} & \multicolumn{3}{c}{LC} \\ \hline
Methods & AG & SD & CL & CL & LI & AR & CL & LI & MA & CL & MA & NI$\downarrow$ \\ \hline
LRRN'23 & 3.72 & 36.41 & {\color[HTML]{3166FF} \textbf{0.19}} & {\color[HTML]{3166FF} \textbf{0.29}} & 1.07 & {\color[HTML]{FE0000} \textbf{0.56}} & 0.19 & 1.24 & 0.23 & 0.14 & {\color[HTML]{FE0000} \textbf{0.17}} & 5.01 \\
CDDF'23 & {\color[HTML]{FE0000} \textbf{5.49}} & {\color[HTML]{3166FF} \textbf{50.87}} & 0.18 & 0.28 & 1.06 & 0.54 & 0.20 & 1.41 & {\color[HTML]{3166FF} \textbf{0.25}} & 0.14 & 0.14 & {\color[HTML]{3166FF} \textbf{4.25}} \\
MMDR'24 & 3.95 & 36.07 & 0.15 & 0.23 & 1.17 & 0.54 & 0.19 & 1.23 & 0.24 & 0.14 & 0.14 & 4.55 \\
SAGE'25 & 4.73 & 43.59 & 0.18 & 0.28 & {\color[HTML]{FE0000} \textbf{1.21}} & 0.54 & 0.20 & 1.17 & 0.20 & 0.15 & 0.14 & 4.48 \\
TITA'25 & 4.93 & 41.68 & 0.18 & 0.29 & 1.17 & 0.54 & 0.18 & 1.45 & 0.24 & 0.14 & 0.14 & 6.38 \\ \hline
Text'24 $\dagger$ & 4.05 & 42.10 & 0.19 & 0.26 & 1.16 & 0.54 & {\color[HTML]{FE0000} \textbf{0.23}} & {\color[HTML]{3166FF} \textbf{1.49}} & 0.23 & {\color[HTML]{3166FF} \textbf{0.16}} & 0.13 & 4.41 \\
DAFu'25 $\dagger$ & {\color[HTML]{3166FF} \textbf{5.19}} & 47.99 & 0.19 & 0.25 & {\color[HTML]{3166FF} \textbf{1.19}} & 0.52 & 0.20 & {\color[HTML]{FE0000} \textbf{1.55}} & 0.24 & 0.13 & 0.10 & {\color[HTML]{FE0000} \textbf{3.17}} \\ \hline
Ours & 4.77 & {\color[HTML]{FE0000} \textbf{55.23}} & {\color[HTML]{FE0000} \textbf{0.20}} & {\color[HTML]{FE0000} \textbf{0.30}} & 1.14 & {\color[HTML]{3166FF} \textbf{0.55}} & {\color[HTML]{3166FF} \textbf{0.21}} & 1.37 & {\color[HTML]{FE0000} \textbf{0.26}} & {\color[HTML]{FE0000} \textbf{0.17}} & {\color[HTML]{3166FF} \textbf{0.15}} & 4.67 \\ \hline
\end{tabular}
}
}
\end{table}

\begin{table}[h]
\renewcommand{\arraystretch}{0.7}
\setlength{\tabcolsep}{0.7mm}{
\caption{Quantitative comparison of degradation-aware fusion tasks. $\dagger$ denotes All-in-One type methods.
}
\label{tab:deg_cmp_2}
\vspace{-0.3cm}
{\fontsize{7pt}{12pt}\selectfont
\begin{tabular}{c|ccc|ccc|ccc|ccc}
\hline
 & \multicolumn{3}{c|}{SR4} & \multicolumn{3}{c|}{SR8} & \multicolumn{3}{c|}{OE+LC} & \multicolumn{3}{c}{LL+LC} \\ \hline
Methods & SF & CL & AR & SF & CL & AR & EN & AG & CL & EN & SD & CL \\ \hline
LRRN'23 & 10.60 & 0.30 & 0.59 & 10.25 & 0.26 & 0.58 & 7.02 & 4.68 & 0.20 & 6.19 & 31.73 & 0.14 \\
CDDF'23 & {\color[HTML]{3166FF} \textbf{14.00}} & {\color[HTML]{3166FF} \textbf{0.32}} & 0.59 & {\color[HTML]{3166FF} \textbf{13.54}} & {\color[HTML]{FE0000} \textbf{0.28}} & {\color[HTML]{3166FF} \textbf{0.58}} & 7.47 & 6.56 & 0.22 & 6.70 & 43.48 & 0.14 \\
MMDR'24 & 9.22 & 0.29 & 0.59 & 9.36 & 0.25 & 0.57 & 7.19 & 4.98 & 0.21 & 6.93 & 37.44 & 0.10 \\
SAGE'25 & 11.81 & 0.26 & {\color[HTML]{3166FF} \textbf{0.59}} & 11.74 & 0.23 & 0.58 & 7.30 & 5.74 & 0.22 & 5.99 & 36.26 & {\color[HTML]{3166FF} \textbf{0.14}} \\
TITA'25 & 12.96 & 0.30 & {\color[HTML]{FE0000} \textbf{0.60}} & 12.59 & 0.26 & 0.59 & 7.43 & 6.13 & 0.20 & 7.06 & 42.66 & 0.11 \\ \hline
Text'24 $\dagger$ & 13.20 & 0.26 & 0.59 & 12.78 & 0.23 & 0.57 & {\color[HTML]{FE0000} \textbf{7.51}} & {\color[HTML]{3166FF} \textbf{6.86}} & {\color[HTML]{3166FF} \textbf{0.23}} & {\color[HTML]{3166FF} \textbf{7.26}} & 49.43 & 0.12 \\
DAFu'25 $\dagger$ & {\color[HTML]{FE0000} \textbf{15.23}} & 0.22 & 0.57 & {\color[HTML]{FE0000} \textbf{15.23}} & 0.24 & 0.58 & 7.47 & 6.37 & 0.20 & 7.24 & {\color[HTML]{3166FF} \textbf{50.84}} & 0.11 \\ \hline
Ours & 11.20 & {\color[HTML]{FE0000} \textbf{0.33}} & {\color[HTML]{FE0000} \textbf{0.60}} & 9.77 & {\color[HTML]{3166FF} \textbf{0.27}} & {\color[HTML]{FE0000} \textbf{0.59}} & {\color[HTML]{3166FF} \textbf{7.48}} & {\color[HTML]{FE0000} \textbf{6.88}} & {\color[HTML]{FE0000} \textbf{0.24}} & {\color[HTML]{FE0000} \textbf{7.42}} & {\color[HTML]{FE0000} \textbf{55.40}} & {\color[HTML]{FE0000} \textbf{0.15}} \\ \hline
\end{tabular}
}
}
\end{table}
Degradation-aware fusion demands models to eliminate degradations and integrate valid source information.

For fair comparison, we use a two-step strategy for degradation-unware methods: SOTA external image restoration followed by image fusion. For Text-IF's limitations on some tasks, \textit{e.g.}, super-resolution and combined degradation, we also use this two-stage approach on these tasks.
In addition, LRRNet \cite{li2023lrrnet}, TITA \cite{hu2025balancing}, SAGE \cite{wu2025every}, MMDRFuse \cite{deng2024mmdrfuse}, and CDDFuse \cite{zhao2023cddfuse} all employ a two-step strategy across tasks.
Task-specific SOTA image restoration models: URetinex (LL) \cite{wu2022uretinex}, SGID-PFF (HZ) \cite{bai2022self}, CoTF (OE) \cite{li2024real}, AirNet (LC) \cite{AirNet}, SwinFuSR (SR) \cite{Arnold_2024_CVPR}. \footnote{For more information including prompts used in experiments, please refer to supplementary materials.}

\subsubsection{Qualitative Comparison}
Fig.\ref{fig:qualitative} qualitatively compare LURE with SOTA methods across degradations, showing LURE's superiority. 

Unlike the existing all-in-One method (DAFusion \cite{wang2025degradation}), LURE benefits from explicit labels as supervisory signals, resulting in richer textural details and less noise in fusion outcomes. Compared to Text-IF \cite{yi2024text}, our method unconstrained by quadruple data format, leverages more high-quality image restoration datasets with real-world scenarios, and yielding more natural images without pre-enhanced sources. Against two-step models like SAGE \cite{wu2025every} or MMDRFuse \cite{deng2024mmdrfuse}, LURE reduces domain shift-related degradation and information loss.

Crucially, consistent with DAFusion, our method inherently addresses cross-modal combined degradations, achieving superior performance without additional combined degradation training datasets. Text-IF, method-limited, struggles with combined degradations, thus needs external restoration, yielding lower quality. Specifically, for OE+LC and LL+LC tasks, our method exhibits superior detail, color vividness, and contrast compared to other approaches. This shows our method can produce high-quality fused images without extra restoration models.

\subsubsection{Quantitative Comparison}
Tab.\ref{tab:deg_cmp_1} and Tab.\ref{tab:deg_cmp_2} quantitatively compares LURE and SOTA methods across degradations. 

LURE generally achieves SOTA performance in 8 tasks. Higher EN, AG, SD, and SF metrics show LURE preserves textural details and clarity. Concurrently, CLIP-IQA (CL), ARNIQA (AR), NUSIQ (NU), MANIQA (MA), LIQE (LI), and NIQE (NI) scores indicate less noise and improved perceptual quality, better aligning with human vision.

\subsection{Multi-modal Semantic Segmentation}
To assess the effectiveness of LURE in high-level tasks, the experiment of multi-modal semantic segmentation is conducted. Segformer-b2 \cite{xie2021segformer} is fine-tuned on MSRS fusion results and evaluated on the test dataset. \footnote{For more details please refer to supplementary material.}

\begin{figure}[!h]
\vspace{-0.55cm}
  \includegraphics[width=0.48\textwidth]{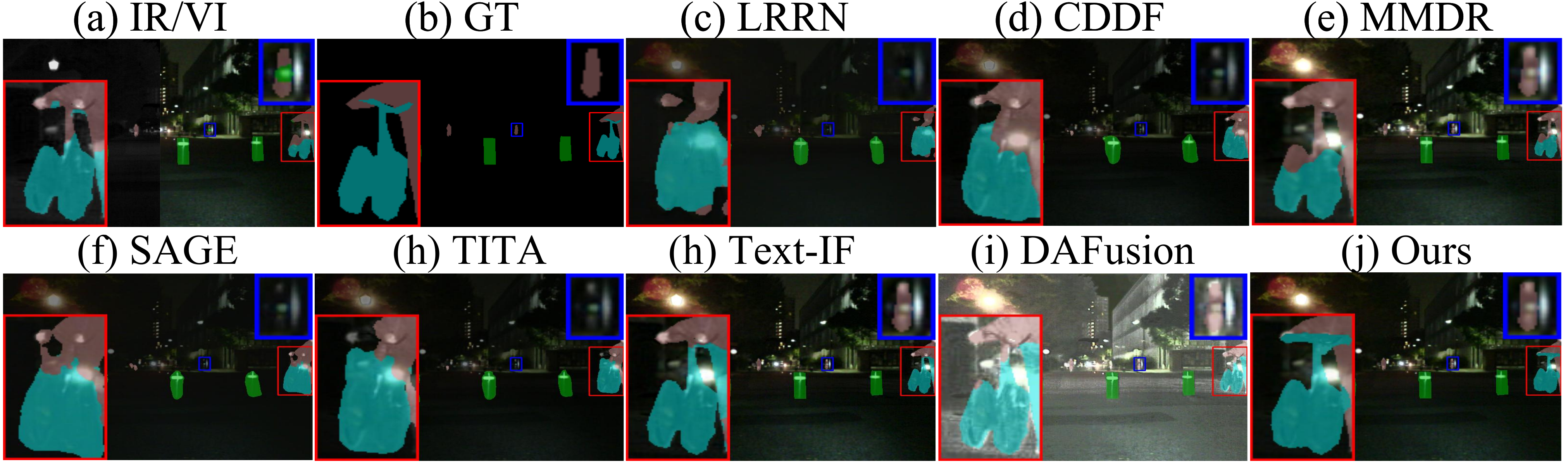}
  \vspace{-1.5\baselineskip}
  \caption{
  Qualitative comparison of multi-modal semantic segmentation on ``00734N" image of MSRS. 
  }
  \label{fig:detec}
\end{figure}

Tab.\ref{tab:detect} presents mIOU scores, showing LURE achieves SOTA segmentation performance, demonstrating visual efficacy on salient objects. Fig.\ref{fig:detec} illustrates LURE preserves more texture details than other approaches, enabling more accurate boundary delineation and improved segmentation.

\begin{table}[h]
\renewcommand{\arraystretch}{0.8}
\vspace{-0.2cm}
\setlength{\tabcolsep}{0.70mm}{
\caption{Quantitative comparison of multi-modal semantic segmentation on MSRS.}
\label{tab:detect}
{\fontsize{7pt}{12pt}\selectfont
\begin{tabular}{c|cccccccccc}
\hline
 & IR & VI & LRRN & CDDF & MMDR & SAGE & FuB & Text & TITA & Ours \\ \hline
mIOU & 63.09 & 66.81 & 66.72 & 67.38 & {\color[HTML]{3166FF} \textbf{69.11}} & 68.64 & 67.33 & 66.26 & 68.37 & {\color[HTML]{FE0000} \textbf{69.25}} \\ \hline
\end{tabular}
}
}
\end{table}

\subsection{Ablation Study}
To validate the effectiveness of our method, several ablation studies are conducted (Tab.\ref{tab:ablation}), including: Single Stage training, Using CC (vs. Cosine Similarity), w/o Inner Residual, TGA$\rightarrow$Gate, w/o Unified Loss and w/o Rule.\footnote{For more detailed qualitative comparisons and hyperparameter ablation studies, please refer to the supplementary material.}

Ablation results show Single Stage training caused learning difficulties due to conflicting losses. TGA$\rightarrow$Gate substitution impairs spatial information perception from images with text guidance.  `w/o Unified Loss' hinders $\mathcal{Z}$ learning, significantly degrading fusion quality across tasks.  

In remaining ablations, `w/o Inner Residual' makes model prone to high-frequency information loss and convergence issues (minor impact). `w/o Rule' slows Stage Two convergence (moderate impact). Conversely, `Using CC', it provides weaker spatial constraints than Cosine Similarity, caused fusion rule transferability issues.
\begin{table}[h]
\renewcommand{\arraystretch}{0.77}
\setlength{\tabcolsep}{0.7mm}{
\caption{Ablation experiment results.}
\label{tab:ablation}
\vspace{-0.25cm}
{\fontsize{7pt}{12pt}\selectfont
\begin{tabular}{c|ccc|ccc|ccc}
\hline
 & \multicolumn{3}{c|}{LC} & \multicolumn{3}{c|}{OE} & \multicolumn{3}{c}{LL+LC} \\ \hline
Configurations & CL & MA & NI $\downarrow$ & CL & LI & MA & EN & SD & CL \\ \hline
Single Stage & 0.168 & 0.135 & {\color[HTML]{FE0000} \textbf{4.397}} & 0.205 & 1.041 & 0.207 & 6.473 & 36.901 & 0.149 \\
Using CC & 0.155 & 0.135 & 4.828 & 0.195 & 1.208 & 0.216 & {\color[HTML]{3166FF} \textbf{7.230}} & 48.035 & 0.134 \\
w/o Inner Residual & 0.165 & 0.138 & 5.100 & {\color[HTML]{3166FF} \textbf{0.206}} & {\color[HTML]{3166FF} \textbf{1.310}} & {\color[HTML]{3166FF} \textbf{0.244}} & 7.056 & 40.099 & {\color[HTML]{3166FF} \textbf{0.149}} \\
TGA$\rightarrow$Gate & 0.163 & 0.130 & 5.000 & 0.180 & 1.080 & 0.200 & 7.184 & 45.001 & 0.143 \\
w/o Unified Loss & {\color[HTML]{3166FF} \textbf{0.169}} & {\color[HTML]{3166FF} \textbf{0.139}} & 5.024 & 0.205 & 1.030 & 0.198 & 7.100 & 45.829 & 0.139 \\
w/o Rule & 0.159 & 0.133 & 5.168 & 0.203 & 1.236 & 0.227 & 7.053 & {\color[HTML]{3166FF} \textbf{48.099}} & 0.136 \\ \hline
Ours & {\color[HTML]{FE0000} \textbf{0.170}} & {\color[HTML]{FE0000} \textbf{0.150}} & {\color[HTML]{3166FF} \textbf{4.671}} & {\color[HTML]{FE0000} \textbf{0.212}} & {\color[HTML]{FE0000} \textbf{1.374}} & {\color[HTML]{FE0000} \textbf{0.259}} & {\color[HTML]{FE0000} \textbf{7.422}} & {\color[HTML]{FE0000} \textbf{55.403}} & {\color[HTML]{FE0000} \textbf{0.152}} \\ \hline
\end{tabular}
}
}
\end{table}

\begin{figure}[!h]
\vspace{-0.45cm}
  \includegraphics[width=0.48\textwidth,height=0.2\textwidth]{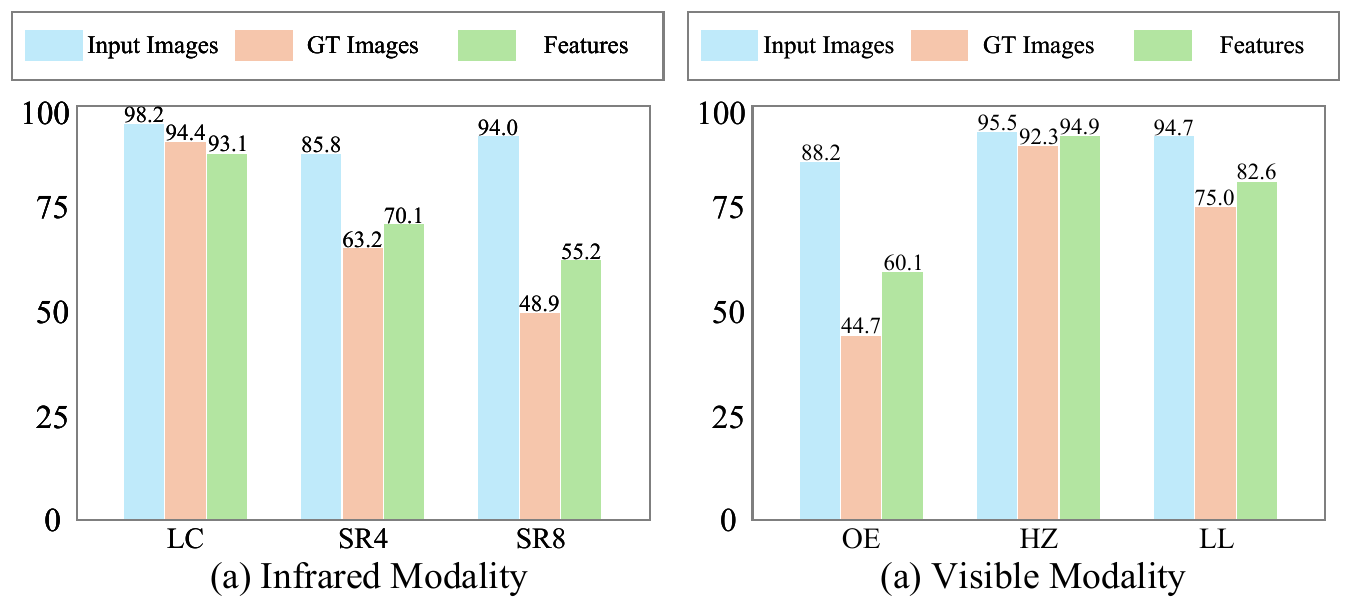}
  \vspace{-1.5\baselineskip}
  \caption{
  Degradation classification accuracy of three classifiers (Input, GT, ULFS) on IR and VI modalities.
  }
  \label{fig:cls}
\end{figure}
\subsection{Validation of Latent Representations}
To verify whether ULFS is successfully learned, we train three ResNet-based classifiers using degraded inputs, ground-truth images, and ULFS features from various restoration datasets, respectively. The inputs contain degradation cues, the ground truths retain only dataset-specific scene distributions, while ULFS features are expected to exclude degradation and preserve scene information. 

We evaluate all classifiers using ACC. As shown in Fig.\ref{fig:cls}, for both IR and VI, the ULFS-based classifier attains accuracy close to the ground-truth one, indicating that degradation information has been effectively removed and the ULFS space has been effectively approximated.

\section{Conclusion}
This paper addresses limitations of prior ADFMs constrained by long-tailed data, dimensionality curse, and synthetic data reliance.  By decoupling modality and quality dimensions at data level and reassociating it at ULFS, our proposed approach mitigates these issues, provides effective supervision, and yields superior fusion performance. Furthermore, an inner residual model enhance spatial perception and detail preservation. Extensive experiments validate our proposed method achieves better performance in vanilla and degradation-aware fusion. Importantly, our method is applicable not only to infrared-visible image fusion but also to other multi-modal image fusion tasks. We believe our approach offers a fresh perspective and inspires future fusion research.

{
    \small
    \bibliographystyle{ieeenat_fullname}
    \bibliography{main}

@String(CVPR= {IEEE Conf. Comput. Vis. Pattern Recog.})

@String(AAAI = {AAAI})

@String(CVPR  = {CVPR})

@article{li2023lrrnet,
  title={Lrrnet: A novel representation learning guided fusion network for infrared and visible images},
  author={Li, Hui and Xu, Tianyang and Wu, Xiao-Jun and Lu, Jiwen and Kittler, Josef},
  journal={IEEE transactions on pattern analysis and machine intelligence},
  volume={45},
  number={9},
  pages={11040--11052},
  year={2023},
  publisher={IEEE}
}

@article{karim2023current,
  title={Current advances and future perspectives of image fusion: A comprehensive review},
  author={Karim, Shahid and Tong, Geng and Li, Jinyang and Qadir, Akeel and Farooq, Umar and Yu, Yiting},
  journal={Information Fusion},
  volume={90},
  pages={185--217},
  year={2023},
  publisher={Elsevier}
}

@article{zhang2025ddbfusion,
  title={DDBFusion: An unified image decomposition and fusion framework based on dual decomposition and B{\'e}zier curves},
  author={Zhang, Zeyang and Li, Hui and Xu, Tianyang and Wu, Xiao-Jun and Kittler, Josef},
  journal={Information Fusion},
  volume={114},
  pages={102655},
  year={2025},
  publisher={Elsevier}
}

@article{zhang2021image,
  title={Image fusion meets deep learning: A survey and perspective},
  author={Zhang, Hao and Xu, Han and Tian, Xin and Jiang, Junjun and Ma, Jiayi},
  journal={Information Fusion},
  volume={76},
  pages={323--336},
  year={2021},
  publisher={Elsevier}
}

@inproceedings{yi2024text,
  title={Text-if: Leveraging semantic text guidance for degradation-aware and interactive image fusion},
  author={Yi, Xunpeng and Xu, Han and Zhang, Hao and Tang, Linfeng and Ma, Jiayi},
  booktitle={Proceedings of the IEEE/CVF Conference on Computer Vision and Pattern Recognition},
  pages={27026--27035},
  year={2024}
}

@inproceedings{liang2022fusion,
  title={Fusion from decomposition: A self-supervised decomposition approach for image fusion},
  author={Liang, Pengwei and Jiang, Junjun and Liu, Xianming and Ma, Jiayi},
  booktitle={European Conference on Computer Vision},
  pages={719--735},
  year={2022},
  organization={Springer}
}

@article{cheng2024fusionbooster,
  title={FusionBooster: A Unified Image Fusion Boosting Paradigm},
  author={Cheng, Chunyang and Xu, Tianyang and Wu, Xiao-Jun and Li, Hui and Li, Xi and Kittler, Josef},
  journal={International Journal of Computer Vision},
  pages={1--18},
  year={2024},
  publisher={Springer}
}

@article{li2020mdlatlrr,
  title={MDLatLRR: A novel decomposition method for infrared and visible image fusion},
  author={Li, Hui and Wu, Xiao-Jun and Kittler, Josef},
  journal={IEEE Transactions on Image Processing},
  volume={29},
  pages={4733--4746},
  year={2020},
  publisher={IEEE}
}

@article{cheng2025textfusion,
  title={TextFusion: Unveiling the power of textual semantics for controllable image fusion},
  author={Cheng, Chunyang and Xu, Tianyang and Wu, Xiao-Jun and Li, Hui and Li, Xi and Tang, Zhangyong and Kittler, Josef},
  journal={Information Fusion},
  volume={117},
  pages={102790},
  year={2025},
  publisher={Elsevier}
}

@inproceedings{zhao2023metafusion,
  title={Metafusion: Infrared and visible image fusion via meta-feature embedding from object detection},
  author={Zhao, Wenda and Xie, Shigeng and Zhao, Fan and He, You and Lu, Huchuan},
  booktitle={Proceedings of the IEEE/CVF Conference on Computer Vision and Pattern Recognition},
  pages={13955--13965},
  year={2023}
}

@article{wang2025degradation,
  title={A degradation-aware guided fusion network for infrared and visible image},
  author={Wang, Xue and Guan, Zheng and Qian, Wenhua and Cao, Jinde and Ma, Runzhuo and Bi, Cong},
  journal={Information Fusion},
  volume={118},
  pages={102931},
  year={2025},
  publisher={Elsevier}
}

@inproceedings{afifi2021learning,
  title={Learning multi-scale photo exposure correction},
  author={Afifi, Mahmoud and Derpanis, Konstantinos G and Ommer, Bjorn and Brown, Michael S},
  booktitle={Proceedings of the IEEE/CVF Conference on Computer Vision and Pattern Recognition},
  pages={9157--9167},
  year={2021}
}

@inproceedings{chen2022simple,
  title={Simple baselines for image restoration},
  author={Chen, Liangyu and Chu, Xiaojie and Zhang, Xiangyu and Sun, Jian},
  booktitle={European conference on computer vision},
  pages={17--33},
  year={2022},
  organization={Springer}
}

@inproceedings{zamir2022restormer,
  title={Restormer: Efficient transformer for high-resolution image restoration},
  author={Zamir, Syed Waqas and Arora, Aditya and Khan, Salman and Hayat, Munawar and Khan, Fahad Shahbaz and Yang, Ming-Hsuan},
  booktitle={Proceedings of the IEEE/CVF conference on computer vision and pattern recognition},
  pages={5728--5739},
  year={2022}
}

@article{chen2023comparative,
  title={A Comparative Study of Image Restoration Networks for General Backbone Network Design}, 
  author={Chen, Xiangyu and Li, Zheyuan and Pu, Yuandong and Liu, Yihao and Zhou, Jiantao and Qiao, Yu and Dong, Chao},
  journal={arXiv preprint arXiv:2310.11881},
  year={2023}
}

@inproceedings{zhao2023ddfm,
  title={DDFM: denoising diffusion model for multi-modality image fusion},
  author={Zhao, Zixiang and Bai, Haowen and Zhu, Yuanzhi and Zhang, Jiangshe and Xu, Shuang and Zhang, Yulun and Zhang, Kai and Meng, Deyu and Timofte, Radu and Van Gool, Luc},
  booktitle={Proceedings of the IEEE/CVF International Conference on Computer Vision},
  pages={8082--8093},
  year={2023}
}

@article{ma2020ganmcc,
  title={GANMcC: A generative adversarial network with multiclassification constraints for infrared and visible image fusion},
  author={Ma, Jiayi and Zhang, Hao and Shao, Zhenfeng and Liang, Pengwei and Xu, Han},
  journal={IEEE Transactions on Instrumentation and Measurement},
  volume={70},
  pages={1--14},
  year={2020},
  publisher={IEEE}
}

@article{ma2019fusiongan,
  title={FusionGAN: A generative adversarial network for infrared and visible image fusion},
  author={Ma, Jiayi and Yu, Wei and Liang, Pengwei and Li, Chang and Jiang, Junjun},
  journal={Information fusion},
  volume={48},
  pages={11--26},
  year={2019},
  publisher={Elsevier}
}

@article{li2025conti,
  title={Conti-Fuse: A novel continuous decomposition-based fusion framework for infrared and visible images},
  author={Li, Hui and Ma, Haolong and Cheng, Chunyang and Shen, Zhongwei and Song, Xiaoning and Wu, Xiao-Jun},
  journal={Information Fusion},
  volume={117},
  pages={102839},
  year={2025},
  publisher={Elsevier}
}

@inproceedings{zhao2023cddfuse,
  title={Cddfuse: Correlation-driven dual-branch feature decomposition for multi-modality image fusion},
  author={Zhao, Zixiang and Bai, Haowen and Zhang, Jiangshe and Zhang, Yulun and Xu, Shuang and Lin, Zudi and Timofte, Radu and Van Gool, Luc},
  booktitle={Proceedings of the IEEE/CVF conference on computer vision and pattern recognition},
  pages={5906--5916},
  year={2023}
}

@article{li2018densefuse,
  title={DenseFuse: A fusion approach to infrared and visible images},
  author={Li, Hui and Wu, Xiao-Jun},
  journal={IEEE Transactions on Image Processing},
  volume={28},
  number={5},
  pages={2614--2623},
  year={2018},
  publisher={IEEE}
}

@article{li2020nestfuse,
  title={NestFuse: An infrared and visible image fusion architecture based on nest connection and spatial/channel attention models},
  author={Li, Hui and Wu, Xiao-Jun and Durrani, Tariq},
  journal={IEEE Transactions on Instrumentation and Measurement},
  volume={69},
  number={12},
  pages={9645--9656},
  year={2020},
  publisher={IEEE}
}

@article{li2024crossfuse,
  title={CrossFuse: A novel cross attention mechanism based infrared and visible image fusion approach},
  author={Li, Hui and Wu, Xiao-Jun},
  journal={Information Fusion},
  volume={103},
  pages={102147},
  year={2024},
  publisher={Elsevier}
}

@article{xu2020u2fusion,
  title={U2Fusion: A unified unsupervised image fusion network},
  author={Xu, Han and Ma, Jiayi and Jiang, Junjun and Guo, Xiaojie and Ling, Haibin},
  journal={IEEE transactions on pattern analysis and machine intelligence},
  volume={44},
  number={1},
  pages={502--518},
  year={2020},
  publisher={IEEE}
}

@inproceedings{conde2024instructir,
  title={Instructir: High-quality image restoration following human instructions},
  author={Conde, Marcos V and Geigle, Gregor and Timofte, Radu},
  booktitle={European Conference on Computer Vision},
  pages={1--21},
  year={2024},
  organization={Springer}
}

@inproceedings{li2024promptcir,
  title={PromptCIR: blind compressed image restoration with prompt learning},
  author={Li, Bingchen and Li, Xin and Lu, Yiting and Feng, Ruoyu and Guo, Mengxi and Zhao, Shijie and Zhang, Li and Chen, Zhibo},
  booktitle={Proceedings of the IEEE/CVF Conference on Computer Vision and Pattern Recognition},
  pages={6442--6452},
  year={2024}
}

@inproceedings{li2022all,
  title={All-in-one image restoration for unknown corruption},
  author={Li, Boyun and Liu, Xiao and Hu, Peng and Wu, Zhongqin and Lv, Jiancheng and Peng, Xi},
  booktitle={Proceedings of the IEEE/CVF conference on computer vision and pattern recognition},
  pages={17452--17462},
  year={2022}
}

@article{wei2018deep,
  title={Deep retinex decomposition for low-light enhancement},
  author={Wei, Chen and Wang, Wenjing and Yang, Wenhan and Liu, Jiaying},
  journal={arXiv preprint arXiv:1808.04560},
  year={2018}
}

@article{li2019benchmarking,

title={Benchmarking Single-Image Dehazing and Beyond},
author={Li, Boyi and Ren, Wenqi and Fu, Dengpan and Tao, Dacheng and Feng, Dan and Zeng, Wenjun and Wang, Zhangyang},
journal={IEEE Transactions on Image Processing},
volume={28},
number={1},
pages={492--505},
year={2019},
publisher={IEEE}
}

@inproceedings{liu2022target,
  title={Target-aware Dual Adversarial Learning and a Multi-scenario Multi-Modality Benchmark to Fuse Infrared and Visible for Object Detection},
  author={Liu, Jinyuan and Fan, Xin and Huang, Zhanbo and Wu, Guanyao and Liu, Risheng and Zhong, Wei and Luo, Zhongxuan},
  booktitle={Proceedings of the IEEE/CVF Conference on Computer Vision and Pattern Recognition},
  pages={5802--5811},
  year={2022}
}

@inproceedings{xu2020aaai,
title={FusionDN: A Unified Densely Connected Network for Image Fusion},
author={Xu, Han and Ma, Jiayi and Le, Zhuliang and Jiang, Junjun and Guo, Xiaojie},
booktitle={proceedings of the Thirty-Fourth AAAI Conference on Artificial Intelligence},
year={2020}
}

@article{Tang2022PIAFusion,
  title={PIAFusion: A progressive infrared and visible image fusion network based on illumination aware},
  author={Tang, Linfeng and Yuan, Jiteng and Zhang, Hao and Jiang, Xingyu and Ma, Jiayi},
  journal={Information Fusion},
  year={2022},
  publisher={Elsevier}
}

@inproceedings{li2023learning,
  title={Learning distortion invariant representation for image restoration from a causality perspective},
  author={Li, Xin and Li, Bingchen and Jin, Xin and Lan, Cuiling and Chen, Zhibo},
  booktitle={Proceedings of the IEEE/CVF Conference on Computer Vision and Pattern Recognition},
  pages={1714--1724},
  year={2023}
}

@inproceedings{poirier2023robust,
  title={Robust unsupervised stylegan image restoration},
  author={Poirier-Ginter, Yohan and Lalonde, Jean-Fran{\c{c}}ois},
  booktitle={Proceedings of the IEEE/CVF Conference on Computer Vision and Pattern Recognition},
  pages={22292--22301},
  year={2023}
}

@inproceedings{devlin2019bert,
  title={Bert: Pre-training of deep bidirectional transformers for language understanding},
  author={Devlin, Jacob and Chang, Ming-Wei and Lee, Kenton and Toutanova, Kristina},
  booktitle={Proceedings of the 2019 conference of the North American chapter of the association for computational linguistics: human language technologies, volume 1 (long and short papers)},
  pages={4171--4186},
  year={2019}
}

@inproceedings{radford2021learning,
  title={Learning transferable visual models from natural language supervision},
  author={Radford, Alec and Kim, Jong Wook and Hallacy, Chris and Ramesh, Aditya and Goh, Gabriel and Agarwal, Sandhini and Sastry, Girish and Askell, Amanda and Mishkin, Pamela and Clark, Jack and others},
  booktitle={International conference on machine learning},
  pages={8748--8763},
  year={2021},
  organization={PmLR}
}

@inproceedings{jia2021llvip,
  title={LLVIP: A visible-infrared paired dataset for low-light vision},
  author={Jia, Xinyu and Zhu, Chuang and Li, Minzhen and Tang, Wenqi and Zhou, Wenli},
  booktitle={Proceedings of the IEEE/CVF International Conference on Computer Vision},
  pages={3496--3504},
  year={2021}
}

@article{oh2024instructir,
  title={Instructir: A benchmark for instruction following of information retrieval models},
  author={Oh, Hanseok and Lee, Hyunji and Ye, Seonghyeon and Shin, Haebin and Jang, Hansol and Jun, Changwook and Seo, Minjoon},
  journal={arXiv preprint arXiv:2402.14334},
  year={2024}
}

@article{tno,
  title={TNO Image Fusion Dataset},
  author={Toet, Alexander},
  journal={figshare},
  year={2024},
  url={https://doi.org/10.6084/m9.figshare.1008029.v2}
}

@article{aslantas2015new,
  title={A new image quality metric for image fusion: The sum of the correlations of differences},
  author={Aslantas, V and Bendes, Emre},
  journal={Aeu-international Journal of electronics and communications},
  volume={69},
  number={12},
  pages={1890--1896},
  year={2015},
  publisher={Elsevier}
}

@article{wang2002universal,
  title={A universal image quality index},
  author={Wang, Zhou and Bovik, Alan C},
  journal={IEEE signal processing letters},
  volume={9},
  number={3},
  pages={81--84},
  year={2002},
  publisher={IEEE}
}

@inproceedings{wang2003multiscale,
  title={Multiscale structural similarity for image quality assessment},
  author={Wang, Zhou and Simoncelli, Eero P and Bovik, Alan C},
  booktitle={The Thrity-Seventh Asilomar Conference on Signals, Systems \& Computers, 2003},
  volume={2},
  pages={1398--1402},
  year={2003},
  organization={Ieee}
}

@article{deshmukh2010image,
  title={Image fusion and image quality assessment of fused images},
  author={Deshmukh, Manjusha and Bhosale, Udhav and others},
  journal={International Journal of Image Processing (IJIP)},
  volume={4},
  number={5},
  pages={484},
  year={2010},
  publisher={Citeseer}
}

@article{roberts2008assessment,
  title={Assessment of image fusion procedures using entropy, image quality, and multispectral classification},
  author={Roberts, J Wesley and Van Aardt, Jan A and Ahmed, Fethi Babikker},
  journal={Journal of Applied Remote Sensing},
  volume={2},
  number={1},
  pages={023522},
  year={2008},
  publisher={SPIE}
}

@inproceedings{wang2023exploring,
  title={Exploring clip for assessing the look and feel of images},
  author={Wang, Jianyi and Chan, Kelvin CK and Loy, Chen Change},
  booktitle={Proceedings of the AAAI conference on artificial intelligence},
  volume={37},
  number={2},
  pages={2555--2563},
  year={2023}
}

@inproceedings{zhang2023liqe,  
  title={Blind Image Quality Assessment via Vision-Language Correspondence: A Multitask Learning Perspective},  
  author={Zhang, Weixia and Zhai, Guangtao and Wei, Ying and Yang, Xiaokang and Ma, Kede},  
  booktitle={IEEE Conference on Computer Vision and Pattern Recognition},  
  pages={14071--14081},
  year={2023}
}

@inproceedings{yang2022maniqa,
  title={MANIQA: Multi-dimension Attention Network for No-Reference Image Quality Assessment},
  author={Yang, Sidi and Wu, Tianhe and Shi, Shuwei and Lao, Shanshan and Gong, Yuan and Cao, Mingdeng and Wang, Jiahao and Yang, Yujiu},
  booktitle={Proceedings of the IEEE/CVF Conference on Computer Vision and Pattern Recognition},
  pages={1191--1200},
  year={2022}
}

@article{mittal2012making,
  title={Making a “completely blind” image quality analyzer},
  author={Mittal, Anish and Soundararajan, Rajiv and Bovik, Alan C},
  journal={IEEE Signal processing letters},
  volume={20},
  number={3},
  pages={209--212},
  year={2012},
  publisher={IEEE}
}

@inproceedings{agnolucci2024arniqa,
  title={Arniqa: Learning distortion manifold for image quality assessment},
  author={Agnolucci, Lorenzo and Galteri, Leonardo and Bertini, Marco and Del Bimbo, Alberto},
  booktitle={Proceedings of the IEEE/CVF Winter Conference on Applications of Computer Vision},
  pages={189--198},
  year={2024}
}

@inproceedings{deng2024mmdrfuse,
  title={MMDRFuse: Distilled Mini-Model with Dynamic Refresh for Multi-Modality Image Fusion},
  author={Deng, Yanglin and Xu, Tianyang and Cheng, Chunyang and Wu, Xiao-Jun and Kittler, Josef},
  booktitle={Proceedings of the 32nd ACM International Conference on Multimedia},
  pages={7326--7335},
  year={2024}
}

@inproceedings{wu2022uretinex,
  title={Uretinex-net: Retinex-based deep unfolding network for low-light image enhancement},
  author={Wu, Wenhui and Weng, Jian and Zhang, Pingping and Wang, Xu and Yang, Wenhan and Jiang, Jianmin},
  booktitle={Proceedings of the IEEE/CVF conference on computer vision and pattern recognition},
  pages={5901--5910},
  year={2022}
}

@article{bai2022self,
    title = {Self-Guided Image Dehazing Using Progressive Feature Fusion},
    author = {Bai, Haoran and Pan, Jinshan and Xiang, Xinguang and Tang, Jinhui},
    journal = {IEEE Transactions on Image Processing},
    volume = {31},
    pages = {1217 - 1229},
    year = {2022},
    publisher = {IEEE}
}

@inproceedings{li2024real,
  title={Real-time exposure correction via collaborative transformations and adaptive sampling},
  author={Li, Ziwen and Zhang, Feng and Cao, Meng and Zhang, Jinpu and Shao, Yuanjie and Wang, Yuehuan and Sang, Nong},
  booktitle={Proceedings of the IEEE/CVF Conference on Computer Vision and Pattern Recognition},
  pages={2984--2994},
  year={2024}
}

@inproceedings{AirNet,
author = {Li, Boyun and Liu, Xiao and Hu, Peng and Wu, Zhongqin and Lv, Jiancheng and Peng, Xi},
title = {{All-In-One Image Restoration for Unknown Corruption}},
booktitle = {IEEE Conference on Computer Vision and Pattern Recognition},
year = {2022},
address = {New Orleans, LA},
month = jun
}

@InProceedings{Arnold_2024_CVPR,
    author    = {Arnold, Cyprien and Jouvet, Philippe and Seoud, Lama},
    title     = {SwinFuSR: An Image Fusion-inspired Model for RGB-guided Thermal Image Super-resolution},
    booktitle = {Proceedings of the IEEE/CVF Conference on Computer Vision and Pattern Recognition (CVPR) Workshops},
    month     = {June},
    year      = {2024},
    pages     = {3027-3036}
}

@inproceedings{zhang2023adding,
  title={Adding conditional control to text-to-image diffusion models},
  author={Zhang, Lvmin and Rao, Anyi and Agrawala, Maneesh},
  booktitle={Proceedings of the IEEE/CVF international conference on computer vision},
  pages={3836--3847},
  year={2023}
}

@article{xie2021segformer,
  title={SegFormer: Simple and efficient design for semantic segmentation with transformers},
  author={Xie, Enze and Wang, Wenhai and Yu, Zhiding and Anandkumar, Anima and Alvarez, Jose M and Luo, Ping},
  journal={Advances in neural information processing systems},
  volume={34},
  pages={12077--12090},
  year={2021}
}

@inproceedings{ke2021musiq,
  title={Musiq: Multi-scale image quality transformer},
  author={Ke, Junjie and Wang, Qifei and Wang, Yilin and Milanfar, Peyman and Yang, Feng},
  booktitle={Proceedings of the IEEE/CVF international conference on computer vision},
  pages={5148--5157},
  year={2021}
}

@article{touvron2023llama,
  title={Llama: Open and efficient foundation language models},
  author={Touvron, Hugo and Lavril, Thibaut and Izacard, Gautier and Martinet, Xavier and Lachaux, Marie-Anne and Lacroix, Timoth{\'e}e and Rozi{\`e}re, Baptiste and Goyal, Naman and Hambro, Eric and Azhar, Faisal and others},
  journal={arXiv preprint arXiv:2302.13971},
  year={2023}
}

@inproceedings{ha2017mfnet,
  title={MFNet: Towards real-time semantic segmentation for autonomous vehicles with multi-spectral scenes},
  author={Ha, Qishen and Watanabe, Kohei and Karasawa, Takumi and Ushiku, Yoshitaka and Harada, Tatsuya},
  booktitle={IEEE/RSJ International Conference on Intelligent Robots and Systems (IROS)},
  pages={5108--5115},
  year={2017},
  organization={IEEE}
}

@inproceedings{hu2025balancing,
  title={Balancing Task-invariant Interaction and Task-specific Adaptation for Unified Image Fusion},
  author={Hu, Xingyu and Jiang, Junjun and Wang, Chenyang and Jiang, Kui and Liu, Xianming and Ma, Jiayi},
  booktitle={Proceedings of the IEEE/CVF International Conference on Computer Vision},
  year={2025}
}

@inproceedings{wu2025every,
  title={Every SAM Drop Counts: Embracing Semantic Priors for Multi-Modality Image Fusion and Beyond},
  author={Wu, Guanyao and Liu, Haoyu and Fu, Hongming and Peng, Yichuan and Liu, Jinyuan and Fan, Xin and Liu, Risheng},
  booktitle={Proceedings of the Computer Vision and Pattern Recognition Conference},
  pages={17882--17891},
  year={2025}
}

@inproceedings{wang2025highlight,
  title={Highlight What You Want: Weakly-Supervised Instance-Level Controllable Infrared-Visible Image Fusion},
  author={Wang, Zeyu and Zhang, Jizheng and Song, Haiyu and Ge, Mingyu and Wang, Jiayu and Duan, Haoran},
  booktitle={Proceedings of the IEEE/CVF International Conference on Computer Vision},
  pages={12637--12647},
  year={2025}
}

@inproceedings{bai2025task,
  title={Task-driven Image Fusion with Learnable Fusion Loss},
  author={Bai, Haowen and Zhang, Jiangshe and Zhao, Zixiang and Wu, Yichen and Deng, Lilun and Cui, Yukun and Feng, Tao and Xu, Shuang},
  booktitle={Proceedings of the Computer Vision and Pattern Recognition Conference},
  pages={7457--7468},
  year={2025}
}
}


\end{document}